\documentclass[11pt]{article}

\usepackage{graphicx}
\usepackage{amsmath}
\usepackage{tikz}
\usepackage{graphicx}
\usepackage{threeparttable}
\usepackage{amsmath}
\usepackage{comment}
\usepackage{layout}
\usepackage{hyperref}
\usepackage{caption}
\usepackage[sort,numbers]{natbib} 
\usetikzlibrary{automata,positioning}

\usepackage{tikz}

\definecolor{lime}{HTML}{A6CE39}
\DeclareRobustCommand{\orcidicon}{
	\begin{tikzpicture}
	\draw[lime, fill=lime] (0,0) 
	circle [radius=0.16] 
	node[white] {{\fontfamily{qag}\selectfont \tiny ID}};
	\draw[white, fill=white] (-0.0625,0.095) 
	circle [radius=0.007];
	\end{tikzpicture}
	\hspace{-2mm}
}

\foreach \x in {A, ..., Z}{\expandafter\xdef\csname orcid\x\endcsname{\noexpand\href{https://orcid.org/\csname orcidauthor\x\endcsname}
			{\noexpand\orcidicon}}
}

\usepackage{xcolor}
\hypersetup{
    colorlinks,
    linkcolor={red!50!black},
    citecolor={blue!50!black},
    urlcolor={blue!80!black}
}

\newcommand{\footremember}[2]{%
    \footnote{#2}
    \newcounter{#1}
    \setcounter{#1}{\value{footnote}}%
}
\newcommand{\footrecall}[1]{%
    \footnotemark[\value{#1}]%
} 

\begin{document}

\title{Best-Answer Prediction in Q\&A Sites Using User Information}
\date{}

\author{%
Rafik Hadfi\orcidA \footremember{ku}{Department of Social Informatics, Kyoto University, Japan}
		\footnote{Correspondence: Rafik Hadfi (rafik.hadfi@i.kyoto-u.ac.jp). Tel.: +81-075-753-4821}%
\and Ahmed Moustafa\footremember{nitech}{Department of Computer Science, Nagoya Institute of Technology, Japan}%
\and Kai Yoshino\footrecall{nitech}%
\and Takayuki Ito\orcidD\footrecall{ku}%
 }

\maketitle

\begin {abstract}
Community Question Answering (CQA) sites have spread and multiplied significantly in recent years. Sites like Reddit, Quora, and Stack Exchange are becoming popular amongst people interested in finding answers to diverse questions. One practical way of finding such answers is automatically predicting the best candidate given existing answers and comments. Many studies were conducted on answer prediction in CQA but with limited focus on using the background information of the questionnaires. We address this limitation using a novel method for predicting the best answers using the questioner's background information and other features, such as the textual content or the relationships with other participants. Our answer classification model was trained using the Stack Exchange dataset and validated using the Area Under the Curve (AUC) metric. The experimental results show that the proposed method complements previous methods by pointing out the importance of the relationships between users, particularly throughout the level of involvement in different communities on Stack Exchange. Furthermore, we point out that there is little overlap between user-relation information and the information represented by the shallow text features and the meta-features, such as time differences.
\end {abstract}


\section{Introduction}

Community Question Answering (CQA) sites play an integral part in our daily lives. People are extensively using sites like Quora or Stack Exchange to find answers to their questions and to share their opinions. As a result, CQA sites have attracted a lot of attention from researchers in social sciences, information retrieval, and machine learning.

An important aspect in CQA platforms is the quality of the content and its relevance to any potential users. Such platforms have several functions to assess the quality of the answers, such as voting, or by using the reputation of its users. Such functions motivate experts to share their knowledge and to answer questions in exchange for more visibility and fame \cite{mamykina2011design}. In this context, the distinguishing function of modern CQA platforms is their ability to select the best answer. Such a function enables questioners to mark one answer as the best reply to their particular question. 

While manually choosing the best answer is sufficient to reach a conclusion, there are many questions for which it is difficult to distinguish the best answer \cite{gkotsis2015acqua}. For instance, a large number of conflicting answers renders the task of finding the best candidate challenging. In this case, the automatic selection of the best answer can reduce the burden on the questioner and stimulate the community. It is therefore worthwhile to automatically search for the best answer, and it has in fact attracted a lot of attention \cite{gkotsis2015acqua,mamykina2011design}.

Despite its practicality, the automatic prediction of the best answer faces a number of challenges. The main challenges are the lack of clear answers to many questions and the large number of participants that might be involved in the discussion. Most of the research in the question answering area does not fully acknowledge those key aspects. For instance, in open-domain question answering, the answers usually exist in open datasets and domains such as Wikipedia. Other similar systems have directly or indirectly indefinite answers in their data or knowledge models \cite{rajpurkar2016squad,cui2019kbqa,zhou2020knowledge}.

Conceptually, CQA systems adopt a simple model where the discussion is centred around a question that is posted by a user and a number of answers provided by the community members \cite{gkotsis2015acqua}. In addition, there is often no clear answer or conclusion to the question, and the questioner has to elect one after seeing all the answers of the users. In this sense, the questioner is a key important factor in determining the best answer regardless of the objective evaluation of the answer.

A common problem with the existing studies on the prediction of the best answer is the assumption that the questioners population is uniform. It is highly unlikely that the questioner shares the same belief and knowledge with the candidate answerers. For instance, the current models are general and assume that the majority of the questioners are set as a reference to the questioners. Such models do not sufficiently handle the background information about the questioner.


The main goal of our work is to address the problem of best answer identification and prediction by combining linguistic, user information, and relationship features. To this end, we set two goals. The first is to increase the performance of automatic best-answer selection based on the proposed set of features. We believe that with the help of AI, future online Q\&A platforms should be able to automatically predict the best answers to questions that lack clear answers. The second goal is the identification of the factors or features that lead to best answers. This goal will contribute to the area of feature engineering in Natural Language Processing (NLP) domains \cite{zheng2018feature}.

In the following, we treat the best answer prediction as a binary classification problem and propose a method that integrates the features used in previous research as well as the background information of the questioner. Our contributions are twofold. Firstly, we focus on the individual features of the questioner as well as the relationships among all the questioners/users. Secondly, we focus on the importance of each questioner/user participation in the community.

The rest of the paper is structured as follows. In the next section, we survey the existing literature on CQA. In Section \ref{chap:propose}, we present the methods. Section \ref{chap:exp} presents our experiments and results. Section \ref{chap:disc} discusses the results. Finally, we conclude and highlight the future directions.

\section{Related Work}

\label{chap:related}

Community Question Answering (CQA) platforms aim at providing high-quality answers to online users. Incomplete or unreliable answers are often returned and the users are usually forced to browse multiple answers in search for the best one. Existing automatic prediction methodologies attempt to find the best answers by using multiple criteria and techniques. Such methodologies are usually based on binary prediction or on ranking the candidate answers. In the following, we cover some of the important methodologies such as ranking, deep learning, expert recommendation, and finally, the user-centred approach that we are adopting in our work.


Machine learning ranking (MLR) has been extensively used in building models to rank web pages, to retrieve online information, and to predict the best answers in CQA systems. MLR methods distinguish between point-wise, pair-wise, and list-wise ranking approaches. The point-wise approach operates on single documents by scoring them according to classification or regression functions obtained from training data. The document scores are then taken as the standard of documents ranking \cite{shah2010evaluating}. The pair-wise approach considers the context of the documents and takes the comparison results as the standard to rank the answers \cite{nie2017data,bian2008finding}. Finally, the list-wise approach does not convert the ranking problem into classification or regression instances, but optimises the ranking results according to predefined evaluation criteria \cite{xu2007adarank}. The authors in \cite{chen2019preference}, for instance, proposed a pair-wise method that ranks the answers based on numerous historical documents. They later applied preference relationships using a deep learning framework. The results showed that the proposed scheme outperformed several state-of-the-art baselines in answer ranking~tasks.


When using machine learning, CQA answer ranking benefits from the recent progress in deep learning methods. For instance, reference \cite{tan2015lstm} combined bidirectional long short term memory (bi-LSTM) with CNN to construct Q\&A text embedding. On the other hand,~reference~\cite{santos2016attentive} proposed an AP-BILSTM model to perform feature weighting of the answers and questions and improve the model performance. Reference \cite{bian2017compare} proposed a novel attention mechanism named Dynamic-Clip Attention that filters the noise in the attention matrices and mines the semantic relevance of word-level vectors. The authors in reference \cite{shen2017inter,tran2018context,tay2018multi} proposed the IWAN, CARNN, and MCAN models respectively by combining attention mechanism and LSTM, where CARNN was an improvement of IWAN. A comprehensive review on deep learning methods applied to answer selection could be found in \cite{lai2018review}.



Other approaches find the best questions by exploiting the expertise of the users in the given topic. For instance, \citep{yang2014tag} extracted the expertise from topics learnt from the content and the tags of the questions and the answers. In practice, the topics of the questions are often general whereas the question tags are specific and therefore more informative. Our method exploits this type of information by focusing on the user expertise.


Other methodologies focus on the features of the users. For instance, reference \cite{gkotsis2015acqua} proposed a method that integrates user-related features such as the reputation captured from features such as answer ratings or age. For instance, the method in \cite{liu2008predicting} used 72 features, including the textual content of the question and the history of the questioner and answerer (number of questions they resolved and the time since their last registration) in order to predict the questioner satisfaction. These methods focus primarily on the answerer's information, and less on the questioner. The authors in \cite{gkotsis2015acqua} did not use the information of the questioner, and the ratio of the feature amount between answerer and questioner used by \cite{liu2008predicting} was 21:4, which indicates a clear bias.


Other than individually focusing on the users, it is possible to look at their social interactions since they are involved in the same thread of discussion. In this context, reference \cite{fang2016community} proposed a framework that encodes social interaction cues in the community in addition to the content of question-answer in order to improve the CQA tasks. This framework used information in heterogeneous social networks. Reference \cite{zhao2017community} developed an asymmetric ranking network learning method with Deep Recurrent Neural Networks by using a heterogeneous asymmetric CQA network in order to consider the authority of the answerers of the question. This heterogeneous asymmetric CQA network is composed by embedding questions, answers, and the users who posted the answers. Despite the usage of the network of features between users, the information about the questioner is neglected.

\section{Methods}
\label{chap:propose}

The proposed method is an extension of \cite{calefato2019empirical} with the addition of new features that lead to significant improvements in the prediction accuracy. The details of the proposed method are explained as follows. 

\subsection{Adopted Discussion Model}
\label{chap:pre:model} 

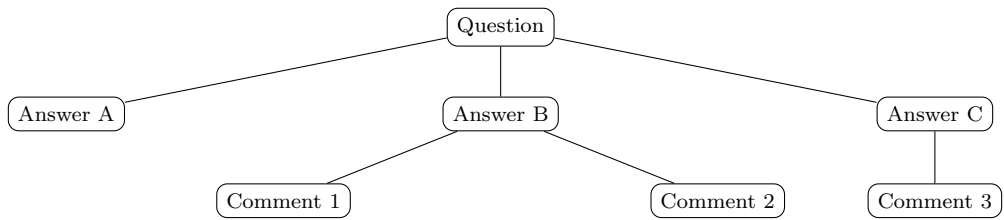
\begin{figure}[h]
\center
\captionsetup{justification=centering}
\begin{tikzpicture}[scale=0.3, sibling distance=50em, level distance=10em,
  every node/.style = {font=\scriptsize, shape=rectangle,
  	 rounded corners, draw, align=center, top color=white}]]
  \node {Question}
    child { node {Answer A} }
    child { node {Answer B} 
        child { node {Comment 1}}
        child { node {Comment 2} } }
    child { node {Answer C}
        child { node {Comment 3} } };
\end{tikzpicture}
  \caption{Question Answering Model}
  \label{fig:discussion-model}
\end{figure}

We start from the hierarchical discussion model used in CQA and illustrated in~Figure~\ref{fig:discussion-model}.

The model starts from the question located at the root node of the tree. Below the question, answers are provided by the users (Answers A, B, and C). Users could also comment below each answer by posting messages (Comments 1, 2, and 3). The question is formulated by the author who could expect people to have divergent points of view. Answers respond to the question and only one question could be chosen as the final solution. Note that comment 1 is expected to be directly related to answer B. Questioning is supposed to end once the best answer is chosen by the questioner. The proposed model is a simplification of the issue-based information system (IBIS) \cite{kunz1970issues} in the sense that both models contain a sequence that starts with a problem followed by a hierarchy of responses and comments.

In this discussion model, the choice of the answer depends mainly on the questioner's choice. In this sense, the author of the question will not necessarily choose the correct answer. Such an answer will later be upvoted or downvoted by the community, which could indicate the absolute correctness of the answer. In our proposed method, incorporating the user features accounts for such discrepancies through the use the histories of votes combined with the acceptance rates. As for the used datasets, questions and answers are usually curated to account for the wrong answers and avoid the inconsistencies that might undermine the predictors.

\subsection{Relevant Features}

The proposed method combines five types of features. We will use these features to train a classification model to predict the best answer to a question. The features of interest are described below.

\begin {itemize} 
\item \textit{Content features} are composed of \textit{shallow} and \textit{textual} features. These features are based on the assumption that the quality of the answer sentence and the degree of association between the question and the answer are important factors to be selected as the best~answer.

\item {User features} are composed of the \textit{answerer}, \textit{questioner}, and \textit{user-relation} features. These user features reflect the intrinsic quality of the users by capturing either their personal history or their interactions with other members of the community. The user features are based on the assumption that different people will conceive their answer differently, which affects the selection of the best answer. Note that the answerer features affect the quality of the users that answered, the questioner features affect the quality of the users who questioned, and the user-relation network reflects the interactions between the users and the effect on the final quality of the answers.
\end {itemize}

In the next section, we cover these two main features in detail.

\subsection{Content Features} 

\subsubsection{Shallow Features} 

Shallow features (\textbf{S}) \cite{gkotsis2014s} do not employ semantic or syntactic parsing such as sentence length \cite{feng2010comparison} or word length \cite{piantadosi2011word}. They are proven to be effective in assessing properties such as the ease of reading, or the usefulness \cite{kincaid1975derivation}. To the best of our knowledge, the most complete approach that uses shallow features is found in \cite{calefato2019empirical}. The work combines the linguistic and meta features that have been emphasized in previous research. In addition, the authors assign ranks after computing numerical values in order to indicate the relative importance. The shallow features follow this adopted approach. The prediction features are listed bellow.

\begin{itemize}
   	 \item [] \textbf{Meta features} 

	\begin{itemize}  
	        \setlength\itemindent{15pt}
                \item  Age 				
                \item Rating score 	
    	 \end{itemize}

  	\item [] \textbf{Linguistic features}
	\begin{itemize}  
	        \setlength\itemindent{15pt}
                \item Length 			
                \item Word count 			
                \item Number of sentences 	
                \item Longest sentence in characters	
                \item Average words per sentence
                \item Average characters per word
                \item Contains hyperlinks or not	
      	   \item Answer count per thread     
         \end{itemize}

  	  \item [] \textbf{Vocabulary features} 
    	 \begin{itemize}  
	        \setlength\itemindent{15pt}
    		\item Normalised Log Likelihood
    		\item Flesch--Kincaid grade
	\end{itemize}

\end{itemize}

Note that for each feature, there exists a ranked version \cite{calefato2019empirical}. In addition to the linguistic features and meta-features, reference \cite{calefato2019empirical} employed two features relevant to the vocabulary used in the questions and answers and are useful in estimating the readability of an answer. These features are the Normalised Log Likelihood and the Flesch--Kincaid~grade.

The Normalised Log Likelihood ($LL_n$) is a probabilistic approach that measures to what extent the lexicon in an answer is distant from the vocabulary used in the whole forum community \cite{calefato2019empirical,gkotsis2014s,gkotsis2015acqua,Pitler08}. In the following, we define $LL_n$ as in (\ref{eq:lln}).

\begin{eqnarray}
LL_n(s) = \frac{1}{U_s} \sum_{w_s} C(w_s) \log(P(w_s|Voc)) 
\label{eq:lln}
\end{eqnarray} 

Given a sentence $s$ in an answer, $P(w_s|Voc)$ is the probability of occurrence of the word $w_s$ according to the background corpus $Voc$, and $C(w_s)$ is the number of times the word $w_s$ occurs in $s$. The normalisation factor $U_s$ is the number of unique words occurring in $s$. While the authors in \cite{calefato2019empirical} argued that $LL_n$ is insignificant, we took the hypothesis that the corpus used in their experiment was too large to be applied individually to each answer. To improve $LL_n$, we propose to use a smaller corpus that is well suited for each answer. While $LL_n$ uses the vocabulary of the whole forum community as a corpus, we propose to use the vocabulary used in the question thread as a corpus. In the case where we allow multiple threads for one single question, all the content of the threads will be compacted into one corpora where messages are temporally ordered below the root node, the question.

The Flesch--Kincaid grade ($F-K$) is a readability metric introduced by \cite{kincaid1975derivation} and is defined as in (\ref{eq:fk}).

\begin{eqnarray}
    \begin{split}
    F-K_p (awps_{p} , asps_p ) = 0.39 \times awps_p + 11.8 \times asps_p - 15.59
     \label{eq:fk}
     \end{split}
\end{eqnarray}

For any given post $p$, the average number of words per sentence ($awps_p$) and the average number of syllables per word ($asps_p$) are calculated and added together. The ranked versions of the rating score and age show the highest importance, indicating that it is important to answer relatively good answers as quickly as possible.

To further represent the shallow textual information, we have also employed {Tag features} extracted from the HyperText Markup Language (HTML) tags. HTML tags are used to visually highlight the question and answers. In the proposed features, the number of specific HTML tags that appear in the answers are calculated and used as features. We distinguish the three following features.

\begin {itemize} 
\item {\textit{Quote}} represents the number of text elements between a ``quote tag'' in an answer. The ``quote tag'' is used to represent quotation from other sources or questions.

\item \textit{Contains} represents the number of quotations originating from the question and used in an answer. A quotation candidate is first selected from the text between ``quote tag'' and then checked whether it is contained in a question or not.

\item \textit{Strong} represents the number of text elements between ``strong tag'' in an answer text. The ``strong tag'' is used to emphasise the text by showing it in bold.
\end {itemize}

Assuming that readability may affect the questioner incentive to choose the best answer, we decided to use the above-mentioned 3 features.
 
\subsubsection{Textual Features}
\label{chap:propose:t}

The textual features (\textbf{T}) aim at estimating the quality and the associations between questions and answers. They are extensively used in \cite{zhou2015answer,zhang2014topic} and can be studied using several analytical tools. To analyse such features, we rely on Latent Dirichlet Allocation (LDA) \cite{blei2003latent} to extract the proportions of topics contained in a document from an input text. These proportions are initially described using probability distributions on questions and answers. Then, information metrics are applied to those distributions to estimate the relevance between question and answer. LDA was trained on the dataset described in Section \ref{chap:data} and tuned with coherence \cite{mimno2011optimizing}. We then employed the model with the highest coherence. We particularly calculated 3 types of probabilistic metrics and 2 types of vectorial metrics between question and answer. Table \ref{tab:lda_feature} describes the features used in the LDA method.

\begin{table}
	\begin{center}
                  \caption{Latent Dirichlet allocation measures}
                  \begin{tabular}{| c | c |} \hline
                  	\textbf{Feature} 	& \textbf{Description} \\ \hline \hline
			$D_{KL}(Q||A)$ 	& KLD from question to answer \\  \hline
                    	$D_{KL}(A||Q)$ 	& KLD from answer to question\\ \hline
                    	$JSD(Q||A)$		& JSD between question and answer  \\ \hline%
                		$R^2(Q,A)$ 		&  $R^2$-score between question and answer \\ \hline%
                		$Cos(Q,A)$ 		& Cosine similarity between question and answer \\ \hline%
                  \end{tabular}
                  \label{tab:lda_feature}
	\end{center}
\end{table}

The measures $D_{KL}(Q||A)$ and $D_{KL}(A||Q)$ are the Kullback-Leibler divergences between question and answer. We treat $D_{KL}(Q||A)$ as the new information from the answer about the question and $D_{KL}(A||Q)$ as the information difference from the answer to the question. We also used Jensen-Shannon divergence (JSD), $R^2$-score, and Cosine similarity, to represent topic similarity between question and answer.

\subsection{User Features}

The features of the users are the most important features in CQA in addition to the textual features \cite{gkotsis2015acqua,liu2008predicting}. Next, we will detail these features and how they contribute to better~predictions.

\subsubsection{Answerer Feature}
\label{chap:propose:af}

The answerer features (AF) are the most emphasised user features. The question and answer history as well as the public profile descriptions are usually available in all major CQA platforms.

Table \ref{tab:user_feature} shows the user features that we used. These features were obtained from the data collected using the official API of Stack Exchange \cite{stexAPI}. Here, {reputation} is a rough measurement of how much the CQA community trusts the user. {reputation} can be earned by posting interesting questions and quality answers. If one user has a reputation of 200 or more, at least on one of Stack Exchange sites, the user will start with 101 reputation as a bonus when he logs into a new site. The features, {bronze}, {silver}, and {gold}, represent the badges obtained by gaining reputation through questions and answers. Badges are difficult to get in the order of {gold}, {silver}, and {bronze}. In addition, the counts {down\_vote\_count}  and {up\_vote\_count} represent the numbers of upvotes and downvotes that the user got from other users. These counts could either raise or lower the reputation of the user.

\begin{table}[h]
\begin{center}
  \caption{User Features}
  \begin{tabular}{| c | c |} \hline
  	\textbf{Feature} 	& \textbf{Description} \\ \hline \hline
	{\it reputation}				& User's reputation \\ \hline
	{\it bronze} 				& Number of bronze badges \\  \hline
	{\it silver} 					& Number of silver badges \\  \hline
	{\it gold} 					& Number of gold badges \\  \hline
	{\it q\_count}, {\it a\_count}	& Posted questions and answers \\ \hline
	{\it up\_vote\_count} & Number of upvotes \\ \hline
	{\it down\_vote\_count} & Number of downvotes \\ \hline
	{\it view\_count}		& Number of views from other users \\ \hline
	{\it accept\_rate}	& Rate of selection as the best answer \\ \hline
  \end{tabular}
  \label{tab:user_feature}
  \end{center}
\end{table}

\subsubsection{Questioner feature}
\label{chap:propose:qf}

We propose to use the questioner features (\textbf{Q}) for the first time in the context of best answer prediction. \textbf{Q} includes the information of the questioner. The questioner is in fact the most important person that can assess if a conclusion is valuable or not. Here, \textbf{Q} relies on features similar to the answerer features but originating from the questioner.

\subsubsection{Difference Feature}

In addition to the \textbf{A} and \textbf{Q} features, we have also used the margins between user features to measure user similarity. The difference is obtained by subtracting the numerical value of the \textbf{A} feature from each numerical value of \textbf{Q} features. This feature is applied only when both \textbf{A} and \textbf{Q} features are used.

\subsubsection{User-Relation Feature}
\label{chap:propose:unfm}

\begin{figure}[h]
\center
\captionsetup{justification=centering}
 \begin{tikzpicture}[>=stealth, node distance=3cm, state/.append style={font=\scriptsize, shape=rectangle, rounded corners, top color=white}] 
	 
            \node[state] (A2) {Questioner};
            \node[state,below right=of A2] (C2) {Other Users};
            \node[state,above right=of C2] (B2) {Answerer};
           
            \draw (A2) edge[->,bend left=15, above] node {$aqSendEdge$} (B2);
            \draw (B2) edge[->,bend left=15, above] node {$qaSendEdge$} (A2);
            
            \draw (C2) edge[<-,bend left=15] node {$qUserSendEdge$} (A2);
            \draw (A2) edge[<-,bend left=15] node {$qUserGetEdge$} (C2);
            
            \draw (C2) edge[<-,bend left=15] node {$aUserSendEdge$} (B2);
            \draw (B2) edge[<-,bend left=15] node {$aUserGetEdge$} (C2);

	\end{tikzpicture}
  \caption{Network of User Relationship Features}
  \label{fig:net_feature}
\end{figure}
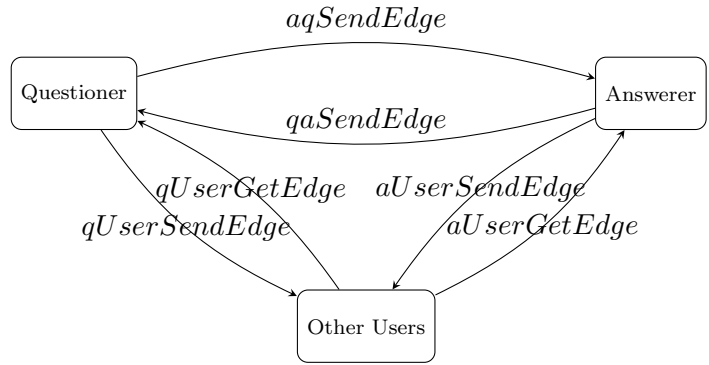

The user-relation (\textbf{UR}) feature was proposed to represent the relationships between users. These graphical features are illustrated on Figure \ref{fig:net_feature} and represent the types of edges when the connections between users are represented as a network. These features are extracted from questions, answers, and comments in the discussion model described in~Section~\ref{chap:pre:model}.

\subsection{Percent Rank Features}
\label{chap:propose:prm}

Another feature we looked at is the percent rank features (PR), and which illustrates feature scaling. In machine learning, feature scaling is useful when the range of values in the feature set is variable. Here, we employed normalisation as a kind of feature scaling. The rank is divided by the number of answers so that the biggest rank value becomes 1 and the smallest rank value approaches 0. The PR method can be applied to all features that possess ranks. In this work, we apply PR to all the features explained in Section \ref{chap:propose}.

\section{Experimental Results}
\label{chap:exp}

Since the goal of our study is to identify the combination of features that could link the threads, meta, linguistic, vocabulary, textual, relationship, and user information all at once, we focused on a dataset that was previously used for a similar task but without looking at the relationships between users \cite{gkotsis2015acqua,gkotsis2014s,cai2011predicting}. Additional evaluations of our metrics with respect to other datasets are required in order to generalize our findings.

\subsection{The Stack Exchange Dataset}
\label{chap:data}

The dataset of our study was collected using the official API of Stack Exchange \cite{stexAPI} and is described in Table \ref{tab:dataset}. Stack Exchange is a group of CQA websites on diverse topics with each site covering a particular theme. Sites in Stack Exchange use a common system in which the best answer is selected. The existence of specific themes as well as the best-answer selection functionality are important criteria for this study. When the themes are separated, we assumed that the users are prompted to participate only on the sites of their themes of interests. In this case, the best-answer selection functionality is necessary to obtain an undisputed conclusion. In sum, the Stack Exchange dataset has the following~advantages.

\begin{enumerate}
  \item The answer rate in all sites is 100\%.
  \item The average number of answers per question in all sites is around 2.2.
  \item The ease of participation whereby people do not need a technical background or a deep knowledge on the topic.
\end{enumerate}

\begin{table}[h]
    \begin{center}
    \caption{Stack Exchange datasets and Q\&A distributions} \label{tab:dataset}
    \begin{threeparttable}
                    \begin{tabular}{| p{1.5cm} | c | c | c |} \hline
                     \textbf{Site} & \textbf{Question} & \textbf{Answer} &
                     			 \textbf{Positive/Negative Ratio} \\ \hline \hline
                      workplace & 11254 & 40722 & $\sim 1:4$ \\
                      english & 34624 & 109337 & $\sim 1:3$ \\ \hline
                    \end{tabular}                    
    \end{threeparttable}
    \end{center}
\end{table}

The positive/negative ratio in Table \ref{tab:dataset} shows that all sites are imbalanced and there are many answers that are not chosen as the best answer. The dataset of each site is composed of questions and answers, including one best-answer for each question. If the dataset presents low answer rates, it should be curated to account for other external features such as the information about the author. This consideration is important when the proposed approach is extended to other CQA platforms.

\subsection{The Experimental Setting}
\label{chap:exp:set}

For the experiment, we prepared and carried out classifications with the four following types of classifiers.

\begin{enumerate}
\setlength{\parskip}{0.1cm} %
\setlength{\itemsep}{0cm} %


\item Support Vector Machines (SVMs), consist in a set of supervised learning approaches used for classification, regression, and outliers detection.

\item Random Forests (RFs), can be used for tasks such as regression and classification by relying on sets of small decision trees with their own local estimators. 

\item Multivariate Adaptive Regression Splines (MARS), are algorithms for complex non-linear regression problems. 

\item Light Gradient Boosting Machine (LGB), is a gradient boosting framework that relies on tree-based learning algorithms. 

\end{enumerate}

The choice of these four classifiers is founded on their proven ability in binary classification problems and particularly when treating major CQA features \cite{calefato2019empirical}. 

Herein, the classifiers were evaluated using a five-fold cross-validation approach with the input dataset being divided into five disjoint partitions. In our case, five folds allowed us to apply the greedy selection procedure in a realistic time. From the five partitions, four training partitions were divided and used to adjust the classifier hyper-parameters, and one partition was used as a test set. To analyse the predictive power of the features, we trained the model on several subsets of features with a greedy selection procedure \cite{molino2016social}. We started by separately testing each feature and picking the performant one, then, we maintained that feature while combining it with all the remaining features to select the best combination. The process is repeated until all features are included. The greedy strategy allows us to find a locally optimal choice at each stage with the hope of finding a global optimum in a reasonable time.

We finally compared the Area Under the Curve (AUC) performance values of the classifiers when using the proposed methods and when using the method of \cite{calefato2019empirical} as a baseline. We employed AUC because of its independence from thresholds.

\subsection{Results}

Table \ref{tab:result:auc} shows the average AUC values for the different classifiers and combinations of features. Letters refer to answerer features (\textbf{A}), questioner features (\textbf{Q}), shallow features (\textbf{S}), textual features (\textbf{T}), percent rank features (\textbf{PR}), and user-relation features (\textbf{UR}). The best combinations of features have the highest AUC value and are shown in bold.

\begin{table}[h]
  \begin{center}
      \caption{Average AUC for feature groups.}
  \begin{threeparttable}[htbp]
    \begin{tabular}{| l |  c | c |  c | c |} \hline
  \textbf{Feature Group}		&   \textbf{SVM}	&   \textbf{RF} 	&   \textbf{MARS}	&   \textbf{LGB} \\ \hline \hline
   S \cite{calefato2019empirical}  & 0.876	& 0.873 	&0.898	&0.945\\ \hline
   {\bf S+PR}  			& 0.866	& 0.869 	&0.917	&0.945\\ 
   T+PR  				& 0.710	& 0.713 	&0.794	&0.824\\ 
   A+PR  				& 0.708	& 0.729 	&0.811	&0.857\\ 
   Q+PR  				& 0.500	& 0.530 	&0.614	&0.628\\ 
   UR+PR  			& 0.664	& 0.683 	&0.774	&0.822\\ \hline
   S +T+PR 			& 0.870	& 0.874 	&0.920	&0.948\\ 
   S +A+PR 			& 0.873	& 0.873 	&0.921	&0.948\\ 
   S +Q+PR 			& 0.872	& 0.874 	&0.921	&0.948\\
   {\bf S +UR+PR} 		& 0.873	& 0.875 	&0.922	&0.950\\ \hline
   S +UR+T+PR 		& 0.870	& 0.875 	&0.919	&0.950\\ 
   {\bf S +UR+A+PR} 	& 0.872	& 0.872 	&0.923	&0.951\\ 
   S +UR+Q+PR 		& 0.873	& 0.874 	&0.922	&0.951\\ \hline
   S +UR+A+Q+PR 		& 0.873	& 0.860 	&0.923	&0.951\\ 
   {\bf S +UR+A+T+PR} 	& 0.870	& 0.867 	&0.924	&0.951\\ \hline \hline
   All					& 0.870	& 0.852	&0.924	&0.952\\ \hline
    \end{tabular}
    \end{threeparttable}
    \label{tab:result:auc}
    \end{center}
\end{table}

The most predictive feature is the feature group {\bf S}. This group contains 22 features that mainly capture the shallow and meta features. These features are very important factors in predicting the quality of the answer. On the other hand, {\bf Q} features alone are expectedly the worst performing group. The nature of the questioner should influence the answer and does not contribute substantially without the answer information. Similar remark could be made for the {\bf UR} group.

The performance of the {\bf S} features has not improved except for MARS when compared to the baseline, which confirms that Normalised Log Likelihood is not important as pointed out and the effect of scaling is small. Interesting results emerged when combining features in pairs. That is, the performance improved mostly by combining the {\bf S} features with the {\bf UR} features that had previously shown the second worst performance. In addition, the {\bf A} features and the {\bf Q} features have a large performance difference when used alone. However, there is no significant performance difference when combined with the {\bf S} features.

These results show that the {\bf S} features, the {\bf A} features, and the {\bf Q} features have overlapping information while the {\bf UR} features have less. The combinations of three or more features made it clear that the {\bf A} features have the greatest effect on the improvement, albeit with a slight difference, and the effects are smaller in the order of {\bf T} and {\bf Q}. Finally, when all the features were used with LGB, the AUC performance reached a maximum of 95.2\%. We used a t-test to verify the difference in AUC values between the case where all features were used in LGB and the baseline. It was confirmed that the difference is significant. For instance, the obtained AUC scores have outperformed the averaged AUC performances of two related baselines. The first one is that of \cite{calefato2019empirical} which focused on thread, meta, linguistic and vocabulary feature ranking applied to Stack Overflow datasets. The second baseline~\cite{gkotsis2014s,gkotsis2015acqua} looked also at the same set features as \cite{calefato2019empirical} but used ten-fold cross-validation in addition to a cross-site training.

We summarise our findings below.

\begin{itemize}
\item Shallow Features are the ones with higher predictive potential compared to other features. 
In particular, the time lag between the answer and the question and the rating of the answer were important, and they showed higher importance than the features proposed in this paper.

\item Among the proposed features, the User-Relation Features, particularly the number of posts by the questioner and the answerer, showed improved performance over other features when combined with the Shallow Features.

\item There was a large difference in performance when used alone, but when combined with other features, Textual Features, Answerer Features, and Questioner Features showed a small difference in performance improvement, which is smaller compared to User-Relation Features.

\end{itemize}

\section{Discussion}
\label{chap:disc}


\subsection{The Proposed Features}

\begin{figure}[h]
\center
\captionsetup{justification=centering}
	 {\includegraphics[scale=0.8]{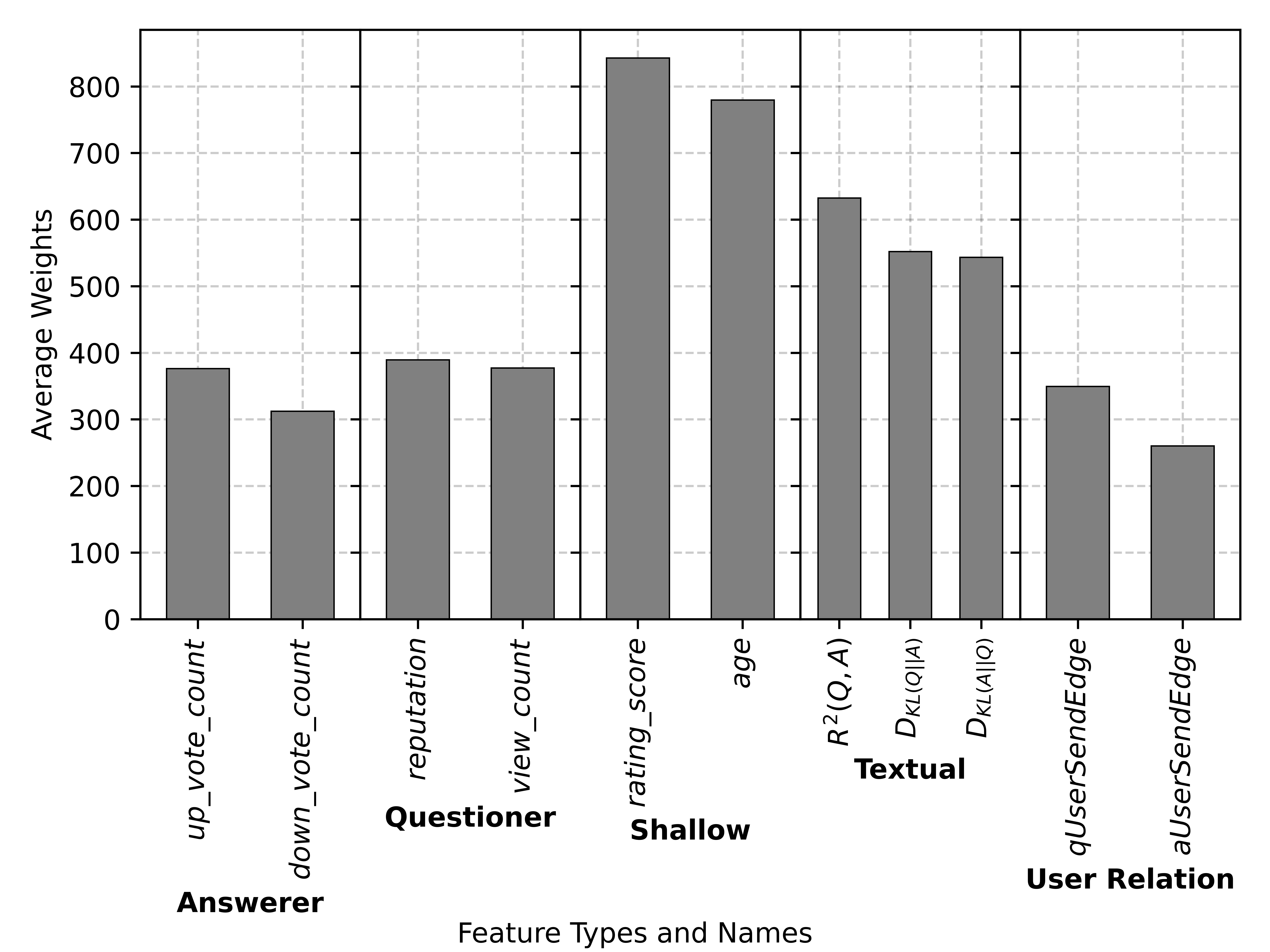}}
      \caption{Important Features of LGB}
    \label{tab:result:weight:pro}
\end{figure}

 Since the LGB showed higher AUC performance than the other classifiers, it could be considered as the most suitable for our prediction problem. Besides the greedy combination of features, to discover which single feature valuations give the best prediction, we compared the average weight of each feature value of the higher-order LGB and its type when all features were used. This comparison of average weights is illustrated in Figure \ref{tab:result:weight:pro}.

The rating score and age of the \textbf{S} features show high weights as it was also found in~\cite{calefato2019empirical}. This evidently indicates the importance of the evaluation of other users and the speed of response. In addition, the proposed features show the weights that follow the baseline features. We consider that these features play a supplementary role in dealing with cases that cannot be explained by the features of the baseline. 

In the following, we describe the consideration for the main values of each feature.

\subsubsection{Textual and Answerer Feature}

The textual feature that had the highest importance was given by $R^2( Q,A )$. Here, $R^2(Q,A)$ was used to calculate topic similarity between question and answer. The difference from $Cos(Q,A) $ is the use of average value. The original purpose of using the average value in $R^2$-score was to measure the relative residual between the predicted value and the measured value. However, the proposed method uses the topic probability distribution for both the predicted value and the measured value. Therefore, the average of predicted value and the measured value is 1 divided by the number of topics. In other words, the $R^2$ score of the proposed features calculates how similar the topic distribution of the answer is to the topic distribution of the question rather than the uniform distribution. It was shown that when compared to a uniform distribution, such similarity is more important than the absolute similarity such as with $Cos(Q,A)$. The features $up\_vote\_count$ and $down\_vote\_count$ showed high weights while $reputation$ did not show higher comparative importance. From these results, it can be said that it is more important for the answerer to consider good and bad evaluations separately than comprehensive evaluations.

\subsubsection{Questioner Features}

Two \textbf{Q} features showed highest weights: The reputation, and the view count of the questioner. The reputation of the questioner was mainly compared with the value near to 101. As we described in Section \ref{chap:data}, 101 is one of the initial reputation values and represents inactive users. {view\_count} was compared with multiple values and no bias was observed. Both { reputation} and {view\_count} are values that increase mainly by the virtue of the questioner's posting to the community. That is, certain or more active participation of the questioner in the community was found to be an important factor for the prediction of best answers.

\subsubsection{User-Relation (UR) Feature}

From the UR features, {qUserSendEdge}, {aUserSendEdge} showed high weights. The feature {qUserSendEdge} was compared to a relatively low numbers (at most 30), and {aUserSendEdge} was compared to high numbers (at least 66 and at most 966). The feature {qUserSendEdge} was compared to a relatively low number, suggesting the existence of some participation of the questioner in the community. That is, certain or more active participation of the questioner in the community was valued with respect to user relationships too. Similarly, {qUserSendEdge} was compared to high numbers, suggesting that it was checking the answerer's quality.  Users with a lot of posts can be said to be actively participating in the community and {aUserSendEdge} can be used to measure the degree of participation independently from the other users when compared to features {reputation}, {up\_vote\_count}, and {down\_vote\_count}.

\section{Conclusions and Future Work}

We explored new features for the prediction of best answers on CQA sites. We started by defining a general discussion model for CQA. Then, we proposed the features that rely on the network information of the questioner and answerer, the background information of the questioner, the sentence features, and the user information of the answerer. We evaluated the performance of the proposed method on a Stack Exchange dataset with a focus on huge groups of CQA websites on diverse topics. We used the AUC value as evaluation index to remove the dependence on thresholds. The experimental results showed that adding the proposed features is superior to the existing method and emphasises the importance of the information about the questioners and answerers, especially their participation in the community.

As a future direction, we will first look at the rating scores. Rating score is an important feature but it is sensitive to change once the best answer is selected. Furthermore, such a score depends highly on other users' valuations. There is therefore a difficulty when using the rating score in real time predictions of best answers. Since the rating score relies on human comprehension, we hypothesise that such a rating could be replaced by a vector space semantic method. Another feature that is worth investigating is the use of emotion tags in CQA and their predictive effect on the correct answers. To account for emotions in CQA, we could either use the existing emotion tags, or classify the emotions directly from the textual content \cite{ekman1993facial}. Finally, we are considering incorporating the proposed metrics in conversational agents that could firstly assess the questions and answers in real time and then react to the users through means of automated facilitation \cite{hadfi2021argumentative,ito2020agent}.



\begin{thebibliography}{999}

\bibitem[Mamykina \em{et~al.}(2011)Mamykina, Manoim, Mittal, Hripcsak, and
  Hartmann]{mamykina2011design}
Mamykina, L.; Manoim, B.; Mittal, M.; Hripcsak, G.; Hartmann, B.
\newblock Design lessons from the fastest q\&a site in the west.
\newblock  In Proceedings of the SIGCHI Conference on Human Factors in Computing
  Systems,  2011;  pp. 2857--2866. 

\bibitem[Gkotsis \em{et~al.}(2015)Gkotsis, Liakata, Pedrinaci, Stepanyan, and
  Domingue]{gkotsis2015acqua}
Gkotsis, G.; Liakata, M.; Pedrinaci, C.; Stepanyan, K.; Domingue, J.
\newblock ACQUA: automated community-based question answering through the
  discretisation of shallow linguistic features.
\newblock {\em J. Web Sci.} {\bf 2015}, {\em 1}, 1--15. 

\bibitem[Rajpurkar \em{et~al.}(2016)Rajpurkar, Zhang, Lopyrev, and
  Liang]{rajpurkar2016squad}
Rajpurkar, P.; Zhang, J.; Lopyrev, K.; Liang, P.
\newblock Squad: 100,000+ questions for machine comprehension of text.
\newblock {\em arXiv} {\bf 2016}, arXiv:1606.05250.

\bibitem[Cui \em{et~al.}(2019)Cui, Xiao, Wang, Song, Hwang, and
  Wang]{cui2019kbqa}
Cui, W.; Xiao, Y.; Wang, H.; Song, Y.; Hwang, S.w.; Wang, W.
\newblock KBQA: learning question answering over QA corpora and knowledge
  bases.
\newblock {\em arXiv} {\bf 2019},  arXiv:1903.02419.

\bibitem[Zhou \em{et~al.}(2020)Zhou, Shi, Huang, and Zhu]{zhou2020knowledge}
Zhou, M.; Shi, Z.; Huang, M.; Zhu, X.
\newblock Knowledge-Aided Open-Domain Question Answering.
\newblock {\em arXiv} {\bf 2020}, arXiv:2006.05244.

\bibitem[Zheng and Casari(2018)]{zheng2018feature}
Zheng, A.; Casari, A.
\newblock {\em Feature Engineering for Machine Learning: Principles and
  Techniques for Data Scientists};  O'Reilly Media, Inc.:~2018. 

\bibitem[Shah and Pomerantz(2010)]{shah2010evaluating}
Shah, C.; Pomerantz, J.
\newblock Evaluating and predicting answer quality in community QA.
\newblock  In Proceedings of the 33rd International ACM SIGIR Conference on
  Research and Development in Information Retrieval,  2010;  pp. 411--418. 

\bibitem[Nie \em{et~al.}(2017)Nie, Wei, Zhang, Wang, Gao, and
  Yang]{nie2017data}
Nie, L.; Wei, X.; Zhang, D.; Wang, X.; Gao, Z.; Yang, Y.
\newblock Data-driven answer selection in community QA systems.
\newblock {\em IEEE Trans. Knowl. Data Eng.} {\bf 2017},
  {\em 29},~1186--1198.

\bibitem[Bian \em{et~al.}(2008)Bian, Liu, Agichtein, and Zha]{bian2008finding}
Bian, J.; Liu, Y.; Agichtein, E.; Zha, H.
\newblock Finding the right facts in the crowd: factoid question answering over
  social media.
\newblock  In   Proceedings of the 17th International Conference on World Wide Web,
  2008;  pp. 467--476. 

\bibitem[Xu and Li(2007)]{xu2007adarank}
Xu, J.; Li, H.
\newblock Adarank: a boosting algorithm for information retrieval.
\newblock In Proceedings of the 30th Annual International ACM SIGIR Conference on
  Research and Development in Information Retrieval, 2007; pp. 391--398. 

\bibitem[Chen \em{et~al.}(2019)Chen, Wang, Lan, and Zheng]{chen2019preference}
Chen, Q.; Wang, J.; Lan, X.; Zheng, N.
\newblock Preference Relationship-Based CrossCMN Scheme for Answer Ranking in
  Community QA.
\newblock  In Proceedings of the   2019 IEEE International Conference on Data Mining (ICDM), Beijing, China, 8--11 November {2019};  pp.~81--90.  

\bibitem[Tan \em{et~al.}(2015)Tan, Santos, Xiang, and Zhou]{tan2015lstm}
Tan, M.; Santos, C.d.; Xiang, B.; Zhou, B.
\newblock Lstm-based deep learning models for non-factoid answer selection.
\newblock {\em arXiv} {\bf 2015}, arXiv:1511.04108.

\bibitem[Santos \em{et~al.}(2016)Santos, Tan, Xiang, and
  Zhou]{santos2016attentive}
Santos, C.d.; Tan, M.; Xiang, B.; Zhou, B.
\newblock Attentive pooling networks.
\newblock {\em arXiv} {\bf 2016}, arXiv:1602.03609.

\bibitem[Bian \em{et~al.}(2017)Bian, Li, Yang, Chen, and Lin]{bian2017compare}
Bian, W.; Li, S.; Yang, Z.; Chen, G.; Lin, Z.
\newblock A compare-aggregate model with dynamic-clip attention for answer
  selection.
\newblock  In Proceedings of the 2017 ACM on Conference on Information and
  Knowledge Management,  2017;  pp. 1987--1990. 

\bibitem[Shen \em{et~al.}(2017)Shen, Yang, and Deng]{shen2017inter}
Shen, G.; Yang, Y.; Deng, Z.H.
\newblock Inter-weighted alignment network for sentence pair modeling.
\newblock In Proceedings of the 2017 Conference on Empirical Methods in Natural
  Language Processing,  {2017}; pp. 1179--1189.

\bibitem[Tran \em{et~al.}(2018)Tran, Lai, Haffari, Zukerman, Bui, and
  Bui]{tran2018context}
Tran, Q.H.; Lai, T.; Haffari, G.; Zukerman, I.; Bui, T.; Bui, H.
\newblock The context-dependent additive recurrent neural net.
\newblock  In Proceedings of the 2018 Conference of the North American Chapter of
  the Association for Computational Linguistics: Human Language Technologies,
  Volume 1 (Long Papers),  2018;  pp. 1274--1283. 

\bibitem[Tay \em{et~al.}(2018)Tay, Tuan, and Hui]{tay2018multi}
Tay, Y.; Tuan, L.A.; Hui, S.C.
\newblock Multi-cast attention networks.
\newblock  In Proceedings of the 24th ACM SIGKDD International Conference on
  Knowledge Discovery \& Data Mining,  2018;  pp. 2299--2308. 


\bibitem[Lai \em{et~al.}(2018)Lai, Bui, and Li]{lai2018review}
Lai, T.; Bui, T.; Li, S.
\newblock A review on deep learning techniques applied to answer selection.
\newblock  In Proceedings of the 27th International Conference on Computational
  Linguistics,  {2018};  pp. 2132--2144. 

\bibitem[Yang and Manandhar(2014)]{yang2014tag}
Yang, B.; Manandhar, S.
\newblock Tag-based expert recommendation in community question answering.
\newblock   In Proceedings of the  2014 IEEE/ACM International Conference on Advances in Social
  Networks Analysis and Mining (ASONAM 2014), Beijing, China, 17--20 August 2014;  pp. 960--963. 

\bibitem[Liu \em{et~al.}(2008)Liu, Bian, and Agichtein]{liu2008predicting}
Liu, Y.; Bian, J.; Agichtein, E.
\newblock Predicting information seeker satisfaction in community question
  answering.
\newblock  In Proceedings of the 31st Annual International ACM SIGIR Conference on
  Research and Development in Information Retrieval, 2008;  pp. 483--490. 

\bibitem[Fang \em{et~al.}(2016)Fang, Wu, Zhao, Duan, Zhuang, and
  Ester]{fang2016community}
Fang, H.; Wu, F.; Zhao, Z.; Duan, X.; Zhuang, Y.; Ester, M.
\newblock Community-based question answering via heterogeneous social network
  learning.
\newblock   In Proceedings of the  Thirtieth AAAI Conference on Artificial Intelligence,  {2016}. 

\bibitem[Zhao \em{et~al.}(2017)Zhao, Lu, Zheng, Cai, He, and
  Zhuang]{zhao2017community}
Zhao, Z.; Lu, H.; Zheng, V.W.; Cai, D.; He, X.; Zhuang, Y.
\newblock Community-based question answering via asymmetric multi-faceted
  ranking network learning.
\newblock   In Proceedings of the Thirty-First AAAI Conference on Artificial Intelligence,  {2017}. 
 
\bibitem[Calefato \em{et~al.}(2019)Calefato, Lanubile, and
  Novielli]{calefato2019empirical}
Calefato, F.; Lanubile, F.; Novielli, N.
\newblock An empirical assessment of best-answer prediction models in technical
  Q\&A sites.
\newblock {\em Empir. Softw. Eng.} {\bf 2019}, {\em 24},~854--901.

\bibitem[Kunz and Rittel(1970)]{kunz1970issues}
Kunz, W.; Rittel, H.W.
\newblock {\em Issues as Elements of Information Systems}; Volume 131, Citeseer: {University Park, PA, USA}, 
  {1970}. 

\bibitem[Gkotsis \em{et~al.}(2014)Gkotsis, Stepanyan, Pedrinaci, Domingue, and
  Liakata]{gkotsis2014s}
Gkotsis, G.; Stepanyan, K.; Pedrinaci, C.; Domingue, J.; Liakata, M.
\newblock It's all in the content: state of the art best answer prediction
  based on discretisation of shallow linguistic features.
\newblock  In Proceedings of the 2014 ACM Conference on Web Science,  {2014}; pp.~202--210. 

\bibitem[Feng \em{et~al.}(2010)Feng, Jansche, Huenerfauth, and
  Elhadad]{feng2010comparison}
Feng, L.; Jansche, M.; Huenerfauth, M.; Elhadad, N.
\newblock \emph{A Comparison of Features for Automatic Readability Assessment}; {{2010}}. 

\bibitem[Piantadosi \em{et~al.}(2011)Piantadosi, Tily, and
  Gibson]{piantadosi2011word}
Piantadosi, S.T.; Tily, H.; Gibson, E.
\newblock Word lengths are optimized for efficient communication.
\newblock {\em Proc. Natl. Acad. Sci. USA} {\bf 2011},
  {\em 108},~3526--3529.

\bibitem[Kincaid \em{et~al.}(1975)Kincaid, Fishburne~Jr, Rogers, and
  Chissom]{kincaid1975derivation}
Kincaid, J.P.; Fishburne, R.P., Jr.; Rogers, R.L.; Chissom, B.S.
\newblock {Derivation of new readability formulas (automated readability index,
  fog count and flesch reading ease formula) for navy enlisted personnel, {1975}}. 

\bibitem[Pitler and Nenkova(2008)]{Pitler08}
Pitler, E.; Nenkova, A.
\newblock Revisiting Readability: A Unified Framework for Predicting Text
  Quality.
\newblock  In \emph{Proceedings of the Conference on Empirical Methods in Natural
  Language Processing}; Association for Computational Linguistics: {Stroudsburg, PA, USA},   {2008};  pp.~186--195. 

\bibitem[Zhou \em{et~al.}(2015)Zhou, Hu, Chen, Tang, and Wang]{zhou2015answer}
Zhou, X.; Hu, B.; Chen, Q.; Tang, B.; Wang, X.
\newblock Answer sequence learning with neural networks for answer selection in
  community question answering.
\newblock {\em arXiv} {\bf 2015}, arXiv:1506.06490.

\bibitem[Zhang \em{et~al.}(2014)Zhang, Liu, Yang, Cao, Zhang, and
  Ji]{zhang2014topic}
Zhang, W.N.; Liu, T.; Yang, Y.; Cao, L.; Zhang, Y.; Ji, R.
\newblock A topic clustering approach to finding similar questions from large
  question and answer archives.
\newblock {\em PloS ONE} {\bf 2014}, {\em 9}, {e71511}. 

\bibitem[Blei \em{et~al.}(2003)Blei, Ng, and Jordan]{blei2003latent}
Blei, D.M.; Ng, A.Y.; Jordan, M.I.
\newblock Latent dirichlet allocation.
\newblock {\em J. Mach. Learn. Res.} {\bf 2003}, {\em
  3},~993--1022.

\bibitem[Mimno \em{et~al.}(2011)Mimno, Wallach, Talley, Leenders, and
  McCallum]{mimno2011optimizing}
Mimno, D.; Wallach, H.M.; Talley, E.; Leenders, M.; McCallum, A.
\newblock Optimizing semantic coherence in topic models.
\newblock  In \emph{Proceedings of the Conference on Empirical Methods in Natural
  Language Processing};  Association for Computational Linguistics: {Stroudsburg, PA, USA}, 2011;  pp.
  262--272.

\bibitem[ste(2019)]{stexAPI}
{Stack Exchange API.} 
\newblock  2019.

\bibitem[Cai and Chakravarthy(2011)]{cai2011predicting}
Cai, Y.; Chakravarthy, S.
\newblock Predicting answer quality in q/a social networks: Using temporal
  features.
\newblock In {\em Arlington: Department of Computer Science and Engineering,
  University of Texas at Arlingtong};  {{2011}}. 

\bibitem[Molino \em{et~al.}(2016)Molino, Aiello, and Lops]{molino2016social}
Molino, P.; Aiello, L.M.; Lops, P.
\newblock Social question answering: Textual, user, and network features for
  best answer prediction.
\newblock {\em ACM Trans. Inf. Syst. (TOIS)} {\bf 2016}, {\em
  35},~1--40.

\bibitem[Ekman(1993)]{ekman1993facial}
Ekman, P.
\newblock Facial expression and emotion.
\newblock {\em Am. Psychol.} {\bf 1993}, {\em 48},~384.

\bibitem[Hadfi \em{et~al.}(2021)Hadfi, Haqbeen, Sahab, and
  Ito]{hadfi2021argumentative}
Hadfi, R.; Haqbeen, J.; Sahab, S.; Ito, T.
\newblock Argumentative Conversational Agents for Online Discussions.
\newblock {\em J. Syst. Sci. Syst. Eng.} {\bf 2021},
  pp. 1--15.

\bibitem[Ito \em{et~al.}(2020)Ito, Hadfi, Haqbeen, Suzuki, Sakai, Kawamura, and
  Yamaguchi]{ito2020agent}
Ito, T.; Hadfi, R.; Haqbeen, J.; Suzuki, S.; Sakai, A.; Kawamura, N.;
  Yamaguchi, N.
\newblock Agent-Based Crowd Discussion Support System and Its Societal
  Experiments.
\newblock In  \emph{International Conference on Practical Applications of Agents and
  Multi-Agent Systems};  Springer: {New~York, NY, USA,} 2020; pp. 430--433. 


\end{thebibliography}
\end{document}